 \newcommand{\ARXIV}{} 

\ifdefined\ARXIV
  \documentclass[letterpaper, 10pt, twocolumn]{article}
  \usepackage[margin=0.75in]{geometry}
  \usepackage{libertine}
  \renewcommand\footnotemark{} 
  \date{}
\else
  \newcommand{\CLASSINPUTtoptextmargin}{17.6mm}
  \newcommand{\CLASSINPUTbottomtextmargin}{25.4mm}
  \newcommand{\CLASSINPUTinnersidemargin}{16.2mm}
  \newcommand{\CLASSINPUToutersidemargin}{16.2mm}
  \documentclass[letterpaper, conference]{IEEEtran}
  \IEEEoverridecommandlockouts
  \usepackage[letterpaper, left=46pt, right=46pt, top=50pt, bottom=76pt]{geometry}
\fi

\usepackage{hyperref}
\hypersetup{colorlinks = true, citecolor = {black}, linkcolor = black}
\usepackage[USenglish]{babel}

\usepackage{amsmath}
\usepackage{amssymb}
\usepackage{babel}
\usepackage{graphicx}
\usepackage{url}

\usepackage{cite}

\title{\LARGE \bf Contact Information Flow and Design of Compliance}

\author{Kevin Haninger, Marcel Radke, Richard Hartisch, J{\"o}rg Kr{\"u}ger
\thanks{Authors affiliated with the Department of Automation at Fraunhofer IPK, Berlin, Germany. Corresponding email: \texttt{kevin.haninger@ipk.fraunhofer.de}. This project has received funding from the European Union's Horizon 2020 research and innovation programme under grant agreement No  820689 — SHERLOCK. }}

\begin{document}

\maketitle
\begin{abstract}
    Identifying changes in contact during contact-rich manipulation can detect task state or errors, enabling improved robustness and autonomy. The ability to detect contact is affected by the mechatronic design of the robot, especially its physical compliance. Established methods can design physical compliance for many aspects of contact performance (e.g. peak contact force, motion/force control bandwidth), but are based on time-invariant dynamic models.  A change in contact mode is a discrete change in coupled robot-environment dynamics, not easily considered in existing design methods.
    
    
    Towards designing robots which can robustly detect changes in contact mode online, this paper investigates how mechatronic design can improve contact estimation, with a focus on the impact of the location and degree of compliance. A design metric of information gain is proposed which measures how much position/force measurements reduce uncertainty in the contact mode estimate. This information gain is developed for fully- and partially-observed systems, as partial observability can arise from joint flexibility in the robot or environmental inertia. Hardware experiments with various compliant setups validate that information gain predicts the speed and certainty with which contact is detected in (i) monitoring of contact-rich assembly and (ii) collision detection.

\end{abstract}

\section{Introduction}
Manipulation aims to achieve a relative arrangement of objects, often with a desired set of contacts between them. In structured environments, this desired contact can be achieved with a certain robot pose, but as robots are applied in less-structured environments, the required robot pose may vary. A robot which can robustly detect contact mode can improve autonomy, for example using online contact detection to switch the controller \cite{bledt2018} or to provide a reward signal to a learning process \cite{vecerik2018}.

\begin{figure}
    \centering
    \includegraphics[width=0.7\columnwidth]{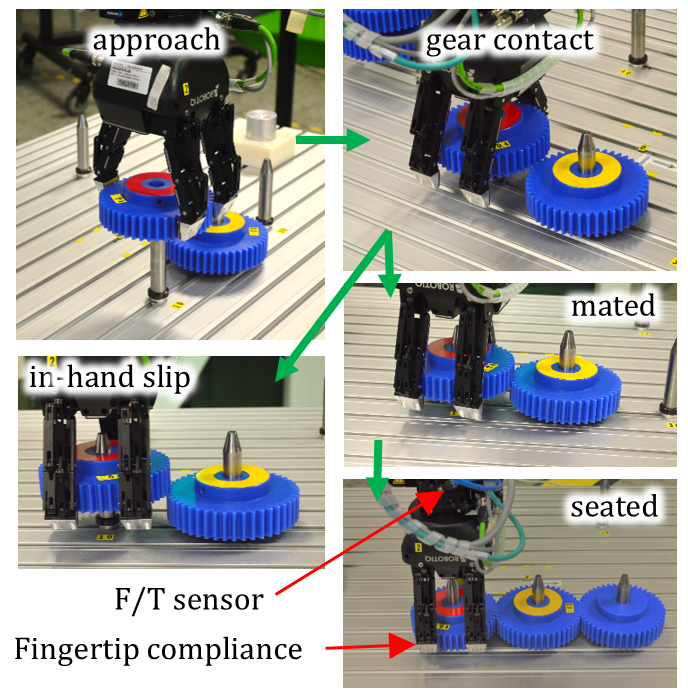}
    \caption{Contact-rich manipulation task, where a discrete collection of contact dynamics occur, corresponding to different task phases. Detecting these changes can help error monitoring like in-hand slip, and the ability to detect these changes is a function of sensor placement and physical dynamics.}
    \label{fig:gear_setup}
\end{figure}

We call a {\it contact mode} a constraint over generalized robot and environment position $q$, i.e. $\phi_i(q) = 0$ for the $i^{\mathrm{th}}$ mode. This is often considered an infinitely stiff kinematic constraint in robotics, either eliminated or enforced with impulses in time-stepping approaches \cite{stewart2000}. Here, we consider a compliant contact model where forces are measured over position $q$ \cite{castro2020, minniti2021a}. Contact can also be described with an inequality $\phi_i(q) \geq 0$ to describe non-penetration of rigid bodies \cite{posa2014, stewart2000}, but we separate the contact conditions into discrete contact modes with equality constraints \cite{minniti2021a}.

Contact mode can be used for task representation and planning. As the contact depends on surface geometry, contact modelling is more common in locomotion where flat surfaces simplify the expression of $\phi(q)$ \cite{posa2014}. In manipulation, when constraints are simple and known in advance, a task can be represented as a sequence of contact modes \cite{fengpan2004}, which can then be used for motion planning \cite{deschutter2007, polverini2019}. Contact geometry can also be demonstrated with special instruments \cite{meeussen2007}. Data-based methods have been used, learning contact geometry on simplified geometries \cite{pfrommer2020}, but not yet demonstrated on robotic manipulation. 

Offline contact models are not robust to variation in environmental objects \cite{rosell2000}, which can be improved with online detection of contact. Estimation of environment dynamics has been established, using a Kalman filter to estimate environment stiffness \cite{rovedaa} or impedance \cite{minniti2021a}. In locomotion, foot contact mode estimation has been proposed in a stochastic framework \cite{bledt2018}, shown to improve robustness to unexpected changes in floor height. In manipulation, discrete contact mode estimation has been proposed through recursive Bayesian estimation \cite{jasim2014, ren2018, haninger2018} and support vector machines \cite{golz2015}. 

The ability to sense contact mode is affected by the hardware design. Hardware design also affects peak collision force \cite{wensing2017, bicchi2004, haninger2019a},  and contact stability \cite{eppinger1992, pratt1995}. The location, degree and structure of compliance can be designed to improve energy efficiency  \cite{bolivar-nieto2021, haddadin2018}, reduce the need for information \cite{haninger2021}, or reduce the problem dimension \cite{hamaya2020}. Compliance can have negative effects, reducing the feasible feedback control gain \cite{tomei1991} and the effectiveness of feedforward control \cite{dellasantina2017}, limiting motion control performance.

The contribution of this paper is a method to design compliance to improve estimation of the contact mode. The estimation of contact mode is established \cite{bledt2018, jasim2014, ren2018, haninger2018}, but the impact of mechatronic design on this estimation is not yet investigated. Analytical rules for `distinguishability' of contact modes have been proposed \cite{debus2005}, but the approach here considers compliance and sensor noise. This paper first introduces the compliant contact models, then the information gain from position and force measurements in the fully- and partially-observed case. The information gain is written in closed form for a linear inertial system. Finally, experiments validate that information gain predicts the certainty and speed of assembly monitoring and collision detection, where the location and degree of compliance is varied. 

\subsubsection*{Notation}
Random variables are typeset in boldface ${\bf n} \sim p(n)$. The normal distribution is denoted $\mathcal{N}(\mu, \Sigma)$, and $\mathcal{N}(x | \mu, \Sigma)$ denotes an evaluation of the Gaussian probability distribution function at $x$. Proportionality $\propto$ indicates a distribution is written without a normalization constant for compactness. The matrix determinant $\mathrm{det}$, and can be ignored when the argument is scalar. A sequence of variables is denoted $q_{1:t} = [q_1, \dots, q_t]$.

\section{Contact and robot models}
This section presents the dynamic model for compliant contact and observation models for position and force.

\subsection{Compliant robot model}
Let the robot and environment be described by generalized position $q\in Q = \mathbb{R}^D$ with velocity $\dot{q}$. Compliance (in the joint, gripper, or environment) can then be modelled by adding to the standard manipulator equations of motion as \cite{spong1987a}
\begin{eqnarray}
M(q)\ddot{q} + \tau(q, \dot{q}) - K(q) = B(q)\tau_m, \label{dynamics}
\end{eqnarray}
where $M(q)\in\mathbb{R}^{D\times D}$ is the inertia matrix, $\tau(q,\dot{q})\in\mathbb{R}^D$ contains the gravitational, Coriolis, and friction forces, spring-like forces $K: \mathbb{R}^D\rightarrow\mathbb{R}^D$, and input matrix $B(q)\in\mathbb{R}^{D\times K}$ modifies motor torque input $\tau_m\in\mathbb{R}^K$.  

This model will be reduced to a two-mass model in the direction of contact motion, but provides the general statement for contact state and observations.

\subsection{Contact State}
Contact is modelled as an equality constraint over total configuration, $\phi(q) = 0$ \cite{stewart2000}. A compliant contact model is used here \cite{minniti2021a, castro2020} instead of impulse-based \cite{stewart2000} -- it is assumed that sufficient compliance is present that collision forces can be represented as a function over configuration $q$. 

While inequality constraints $\phi(q)\geq 0$ can capture non-penetration constraints, thereby modelling both free-space and constrained dynamics in a unified way \cite{posa2014}, the approach here instead uses a collection of equality constraints, only one of which is active at once. Therefore, the contact state is then a discrete variable $n\in[1,\dots,N]$, where each contact state $n$ has a corresponding constraint $\phi_n(q)=0$.  

\subsection{Observation model}
Denote a position measurement at time step $t$ which measures some of the robot/environment positions $\theta_t\in\Theta\subset Q$, representing, e.g. that motor positions are measured, but the position of links or environmental inertias are not.  

Let the measured torques/forces $f_t \in \mathbb{R}^L$ be a subset of the forces on compliant elements,
\begin{equation}
    {\bf f}_t = K^n_f(q_t)+{\bf \epsilon}_t \label{force_measurement} 
\end{equation}
where ${\bf \epsilon}_t\sim\mathcal{N}(0, \Sigma_f)$ is i.i.d. noise capturing electrical or quantisation noise, and $K^n_f: Q\rightarrow \mathbb{R}^L$ the effective sensor stiffness for the mode $n$. Therefore
\begin{equation}
    p(f_t | n, q_t) = \mathcal{N}(K^n_f(q_t), \Sigma_f). \label{force_measurement_dist}
\end{equation}

As force/torque sensors measure displacement - either strain or bulk displacement  \cite{kashiri2017} - practical force measurements can be described with \eqref{force_measurement}.

\section{Contact Information Gain}
Bayesian inference is used to recursively infer contact mode given a prior belief of contact mode, position and force measurements.  The prior distribution can reflect a priori assumptions about the environment, e.g., that a contact occurs at a certain location, or it could be a belief informed by other sensors. 

\subsection{Fully Observed}
When the system is fully observed, i.e. $\Theta = Q$, a prior belief $p({ n})$ is updated by a measurement $f,\, \theta$ as
\begin{eqnarray}
    p({ n} | f, \theta) = \frac{p(f|n, \theta)p({n})}{\sum_{m=1}^N p(f|m,\theta)p(m)},
\end{eqnarray}
where the denominator is a normalization constant.

In the fully observed case, multiple measurements are independent as $\epsilon_t$ in \eqref{force_measurement} is i.i.d., and the full posterior is
\begin{eqnarray}
   p(n | f_{1:t}, \theta_{1:t}) \propto p(n) \prod_{i=1}^t p(f_i | n, \theta_{i}).
\end{eqnarray}
This can be updated recursively as a hidden Markov model \cite{thrun2002}.

\subsection{Partially observed}
In the partially observed case, some positions are not directly measured. This can arise from joint flexibility (motor position is measured, link position is not), or from inertia in the environment. In this case, there are force contributions from the unmeasured inertias which will appear with the forces resulting from the contact mode.

We denote measured position $\theta_t$ and unmeasured position $\overline{q}_t$, so $q_t = [\theta_t^T, \overline{q}_t^T]^T$, making the force observation model $p(f_t | \overline{q}_t,\theta_t,n)$. A dynamic model such as a discretized version of \eqref{dynamics} can be used for inference, where we assume the dynamics $p(\overline{q}_{t+1}|n, \overline{q}_t,\theta_t)$ are known (these are derived for linear inertial system in Section \ref{part_obs_inertial}).  With these dynamics, a standard recursive estimate can be written \cite[Ch. 2.4]{thrun2002}
\begin{equation}
\begin{split}
   p(\overline{q}_{t} | n, \theta_{1:t-1}, f_{1:t-1}&) \\
   \propto \int p(\overline{q}_1)\prod_{i=1}^{t-1} & p(\overline{q}_{i\texttt{+}1} | n, \overline{q}_{i}, \theta_{i})p(f_i | n, \overline{q}_i, \theta_i)\overline{q}_{1:t\texttt{-}1}, \label{state_est}
\end{split}
\end{equation}
where $p(\overline{q}_1)$ is a distribution of initial position ${\bf \overline{q}}_1$. To calculate \eqref{state_est}, a Kalman filter can be used in the linear case \cite[Ch. 3.2]{thrun2002}, and in the nonlinear flexible-joint robot case, various observers of link-side position/velocity are established \cite{haddadin2017}. 

The observation model \eqref{force_measurement_dist} then needs to be marginalized over ${\bf \overline{q}}_t$ as
\begin{equation}
\begin{split}
   p(f_{t} | n, f_{1:t\texttt{-}1}, \theta_{1:t}) \\ \texttt{=} \int  p(f_{t}  | n, \overline{q}_{t}, & \theta_{t}) p(\overline{q}_{t} | f_{1:t\texttt{-}1}, \theta_{1:t\texttt{-}1},n) d\overline{q}_{t}.
\end{split}
\end{equation}

We assume the force observations \eqref{force_measurement} are linear, i.e. $K^n_f(q_t) = K^n_f(q_t - q_0^n)$ where $K^n_f\in\mathbb{R}^{L\times D}$ and $q_0^n$ is the rest position of the spring.  Denote the posterior belief of $p(q_t | n, f_{1:t-1}, \theta_{1:t-1}) = \mathcal{N}(\mu^n_q, \Sigma^n_{q})$, then the force measurement is distributed as
\begin{equation}
    p(f_t | n, f_{1:t\texttt{-}1}, \theta_{1:t\texttt{-}1})  = \mathcal{N}(K_f^n(\mu^n_q\mathtt{-}q^n_0), K_f^{n}\Sigma_q^n K_f^{n,T} \mathtt{+} \Sigma_f).
\end{equation}
Note that the covariance in ${\bf f}_t$ now depends on the uncertainty in the position $\Sigma_q^n$, the stiffness $K_f^n$, and force sensor noise $\Sigma_f$. 

\subsection{Information gain}
The information gain is proposed as the reduction in uncertainty from $p(n)$ to $p(n|f,\theta)$, i.e. how much a measurement reduces belief entropy. Recall that entropy measures the uncertainty of a discrete random variable $H({\bf n}) = -\sum_{n=1}^N p(n)\ln(p(n))$ and differential entropy over continuous random variables $h({\bf x}) = -\int p(x) \ln p(x) dx$ \cite[Ch. 2]{cover1999}, we define the information gain as
\begin{eqnarray}
    \Delta_I = H({\bf n}) - H({\bf n} | {\bf f} , \theta). \label{info_gain}
\end{eqnarray}
A larger $\Delta_I$ means that the measurement brings more certainty.

\begin{figure}
\centering
\includegraphics[width = 0.7\columnwidth]{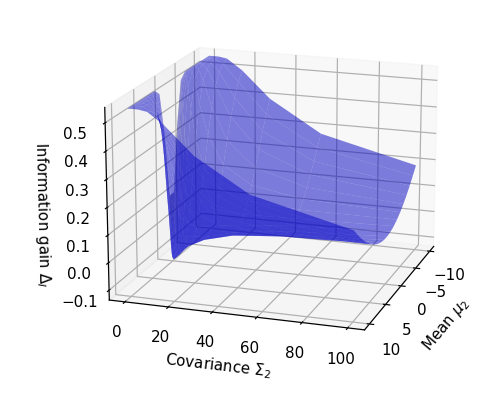}
\caption{Information gain for single observation with two modes, where $\mu_1 = 0$, $\Sigma_1 = 1$.  As the differences between the two distributions increase, the information gain increases, and the mode can be more accurately inferred. \label{info_gain_demo}}
\end{figure}

Denoting a belief vector $b\in \mathbb{R}^N$, where $p(n=m) = b_m$,  $b_m > 0$, $\Vert b\Vert_1=1$,  and the distribution of $\bf f$ per mode as 
\begin{equation}
    p({ f}|n,\theta) = \mathcal{N}({ f}|\mu_n, \Sigma_n),\label{per_mode}
\end{equation}
the following relations hold:
\begin{align}
    H({\bf n}|{\bf f},\theta) = & H({\bf n})-h({\bf f}|\theta)+h({\bf f}|{\bf n},\theta), \label{ent_relation} \\
    h({\bf f}|{\bf n},\theta) = & \frac{L}{2}\ln2\pi e+\sum_{n=1}^N\frac{b_{n}}{2}\ln\det\Sigma_{n}, \label{ent_conditional} \\
    h({\bf f}|\theta) \geq & -\sum_{n=1}^N b_{n}\ln\left(\sum_{m=1}^Nb_m\mathcal{N}\left(\mu_m|\mu_n, \Sigma_n\mathtt{+}\Sigma_m\right) \right),\,\,\,\,\,\,\label{ent_marginal}
\end{align}
where \eqref{ent_relation} results from writing $I({\bf f}, {\bf n}|\theta)$ two ways, $h({\bf f}|\theta) - h({\bf f}|{\bf n},\theta) = H({\bf n} |\theta) - H({\bf n}|{\bf f},\theta)$ \cite{kolchinsky2017},  \eqref{ent_conditional} is standard multivariate Gaussian entropy with the expectation taken over $\bf n$ \cite{cover1999}, and \eqref{ent_marginal} is directly Theorem 2 of \cite{huber2008}.  Then, $\Delta_I$ can be lower bounded as $\Delta_I\geq \underline{\Delta}_I$ where
\begin{eqnarray}
\begin{split}
\underline{\Delta}_I = & -\sum_{n=1}^N b_n \ln \sum_{m=1}^N b_m\mathcal{N}(\mu_n | \mu_m, \Sigma_n+\Sigma_m) \\ & - \frac{L}{2}\ln2\pi e-\sum_{n=1}^N\frac{b_{n}}{2}\ln\det\Sigma_{n}.  \label{delta_i}
\end{split}
\end{eqnarray}
To demonstrate, the information gain bound $\underline{\Delta}_i$ with two modes $N=2$, one dimension $L=1$, and a flat prior belief $b_n = 0.5$ is shown in Figure \ref{info_gain_demo}, where $\mu_1=0$, $\Sigma_1 = 1$, and $\mu_2,\,\Sigma_2$ are varied. The minimum is reached when $\mu_2 = 0$, $\Sigma_2 = 1$, where the two distributions are identical. As $\Sigma_2$ decreases, the information gain increases, as well as when $|\mu_2 - \mu_1|$ increases. Note that the information gain saturates - the maximum reduction in uncertainty is the prior uncertainty $H({\bf n})$. 

Interpreting this for the contact mode, when the mean force expected from two contacts vary widely, a force measurement should provide more information gain and the mode uncertainty should decrease quickly.  If the mean of the force expected in the two modes is the same, there may be some information gain as long as the covariance is different. However, if the mean and covariance of two contact mode forces are the same, there is no reduction in mode uncertainty.

In the partially-observed case, $\underline{\Delta}_I = H({\bf n})-H({\bf n}|{\bf f_t},f_{1:t-1}, \theta_{1:t})$, modifying the force distribution \eqref{per_mode}.

\section{Information gain in inertial systems}

\begin{figure}
\centering
\includegraphics[width = .9\columnwidth]{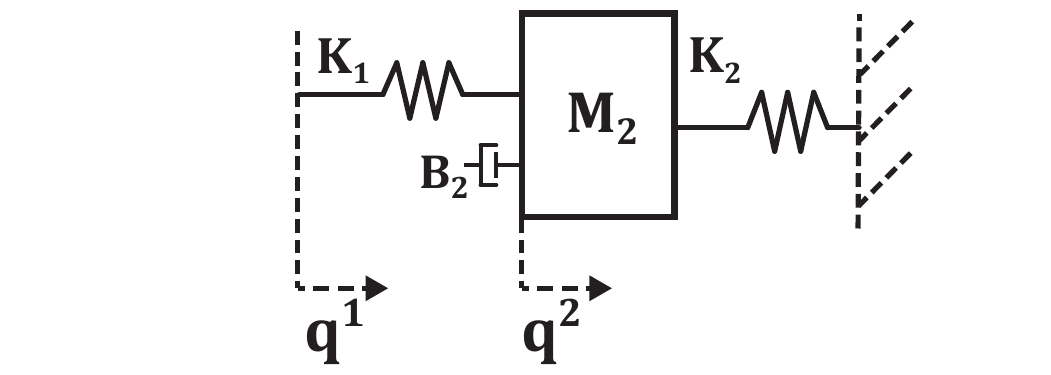}
\caption{Linear inertial model, where $q^1$ is measured robot or motor position, and $q^2$ possibly unmeasured. \label{two_mass}}
\end{figure}

In this section, information gain is considered on a linear, inertial system as seen in Figure \ref{two_mass}. We consider that force is measured as $f_t = K_1(q^2_t-q^1_t)+\epsilon_t$, where $\epsilon_t \sim \mathcal{N}(0,\Sigma_f)$.

Consider two modes, free space where $K_1=0$, and contact with $M_2$ where $K_1> 0$. This gives
\begin{equation}
    \begin{split}
        p(f | n=1) = & \,\,\mathcal{N}(0, \Sigma_f) \\
        p(f | n=2, q^1, q^2)  = & \,\,\mathcal{N}(K_1(q^1-q^2), \Sigma_f).  \label{force_obs}
    \end{split}
\end{equation}
 
\subsection{Fully observed}
In the fully observed case, the positions of $q^1,q^2$ are known and there are no unmeasured inertias. In other words, the robot is stiff, the environment has no inertias, and the contact force model \eqref{force_measurement_dist} can be directly evaluated. If we consider the information gain at a flat prior $b_n = 0.5$, \eqref{delta_i} can be used with \eqref{force_obs} to find
\begin{align}
\underline{\Delta}_I(q^1, q^2)   = & -\ln \left(\mathcal{N}(0|K_1(q^1-q^2), 2\Sigma_f)\right)\nonumber \\  &- \frac{1}{2}\ln2\pi e-\frac{1}{2}\ln\Sigma_{f}\\
  = & \frac{1}{2}K_1^2\left(q^1-q^2\right)^{2}(2\Sigma_{f})^{-1}-\frac{1}{2}(1-\ln2), \label{info_gain_full} 
\end{align}
using $\mathcal{N}(\mu_n|\mu_m, \Sigma_n+\Sigma_m) = \mathcal{N}(\mu_m|\mu_n, \Sigma_n+\Sigma_m)$. 

Here, we note that the information gain increases as $K_1$ does, $|q^1-q^2|$ grows, or as sensor noise $\Sigma_f$ decreases. When we have a stiff environment, deeper contact, or a less noisy sensor, the ability to estimate contact mode increases.

\subsection{Partially observed \label{part_obs_inertial}}
Here we consider when $q^2$ is not observed, e.g. $q^1$ is motor position and $q^2$ is the position of an environment or link inertia. The contact force model \eqref{force_measurement_dist} cannot be directly evaluated, first $q^2$ must be estimated, increasing uncertainty. As time is relevant in the partially observed case, we denote $q^1_t$ and $q^2_t$ as the values at time step $t$, and consider that $q^1_t$ is measured, along with force as in \eqref{force_obs}.  The first-order discretized system dynamics can be written as
\begin{equation}
\begin{split}
\left[\begin{array}{c}
q^2_{t+1}\\
\dot{q}^2_{t+1}
\end{array}\right] & \approx\left[\begin{array}{cc}
1 & T_{s}\\
-T_s\frac{K_{2}+K_1}{M_2} & 1-T_{s}\frac{B_{2}}{M_2}
\end{array}\right]\left[\begin{array}{c} q^2_t\\ \dot{q}^2_t \end{array}\right]\\
& +\left[\begin{array}{c} 0\\ T_sM_{2}^{-1}K_1  \end{array}\right]q^1_t + \left[\begin{array}{c} 0\\ M_{2}^{-1} \end{array}\right]{\bf w}_t\label{two_mass_dyn} \\
&  =  A\left[\begin{array}{c} q^2_{t+1}\\ \dot{q}^2_{t+1} \end{array}\right] + Bq^1_t + B_w {\bf w_t} 
\end{split} 
\end{equation} 
where ${\bf w}_t \sim \mathcal{N}(0, \Sigma_w)$ is i.i.d. process noise which is assumed to be a force disturbance, and thus multiplied by $M_2^{-1}$. 

A Kalman estimator can be written, where, for simplicity, we consider the covariance of $p(q^2_t|f_{1:t},q^1_{1:t})$ as $t\rightarrow \infty$, which is given by the steady-state covariance of  $P$, which satisfies the discrete algebraic Riccati equation of 
\begin{equation}
    P = APA^T-APC^T(CPC^T+\Sigma_f)^{-1}CPA^T+B_w\Sigma_wB_w^T,
\end{equation}
where $A$, $B$ and $B_w$ are from \eqref{two_mass_dyn} and $C=[K_1, 0]$ \cite{thrun2002}. 

While the steady-state covariance $P$ is not necessarily the covariance during transient phases (i.e. when starting from an uncertain initial condition), it has a simple closed-form solution which is differentiable \cite{kao2020}. 

With this assumption, $p(q^2_t | f_{1:t-1},q^1_{1:t-1}) \sim \mathcal{N}(\mu^2_t, P^+)$, where $P^+= APA^T+B_w\Sigma_wB_w^T$ and the posterior distribution on ${\bf f}_t$ can be written as:
\begin{equation}
\begin{split}
p(f_{t}|n=2,q^1_{1:t},f_{1:t-1}) = \qquad \qquad \qquad \qquad \qquad  \\\mathcal{N}(K_1(\mu^2_{t}-q^1_t), K_1P^+K_1^T+\Sigma_f), \label{partially_obs}
\end{split}
\end{equation}
updating the contact force distribution in \eqref{force_obs}.

\section{Experimental Results}
We test information gain in experiments with industrial robots (7 and 60 Kg payload) and varying compliant conditions. An F/T sensor is integrated at the flange and measured at $1250$ Hz, and a standard admittance control and collision detector implemented, see \cite{haninger2022} for details. When collision is detected based on a force threshold, the position commanded to the robot is held to the last position before collision was detected.  The F/T sensors used in all setups have similar noise properties, and $\Sigma_f = 1.25$ is used for all conditions. 

The experimental data and analysis code is available at \url{https://owncloud.fraunhofer.de/index.php/s/ovtOlGRb7o7b9Fb}.

\subsection{Monitoring Assembly}
We consider an admittance-controlled gear assembly task from \cite{haninger2018}, where the gear assembly has multiple stages, as seen in Figure \ref{fig:gear_setup}. One failure mode is in-hand slip, which may occur during the gear mating process. We modify the setup by introducing additional foam at the fingertips, between the gripper and gear, and compare the ability to infer the task mode.  The parameters of the estimator are available with the experimental data in the above mentioned cloud folder. 

\begin{figure}
\centering
\includegraphics[width = 0.8\columnwidth]{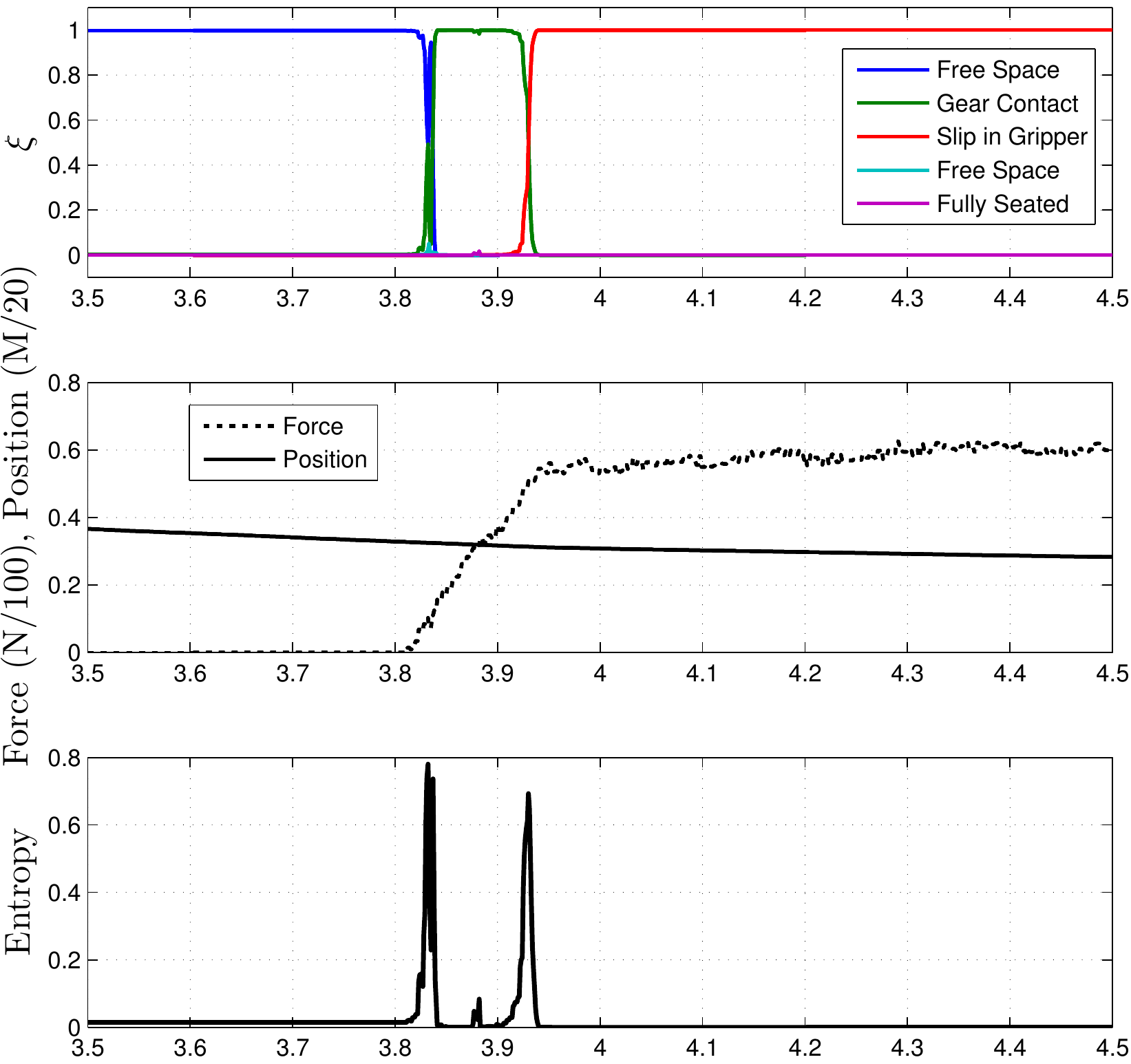}
\caption{Detail of transition in mode estimate, with estimated contact state over time (top), force/position (middle) and belief entropy at each time step (bottom). \label{fig:entropy_detail}}
    \vspace{-4mm}
\end{figure}

\begin{figure}
\centering
\includegraphics[width = 0.8\columnwidth]{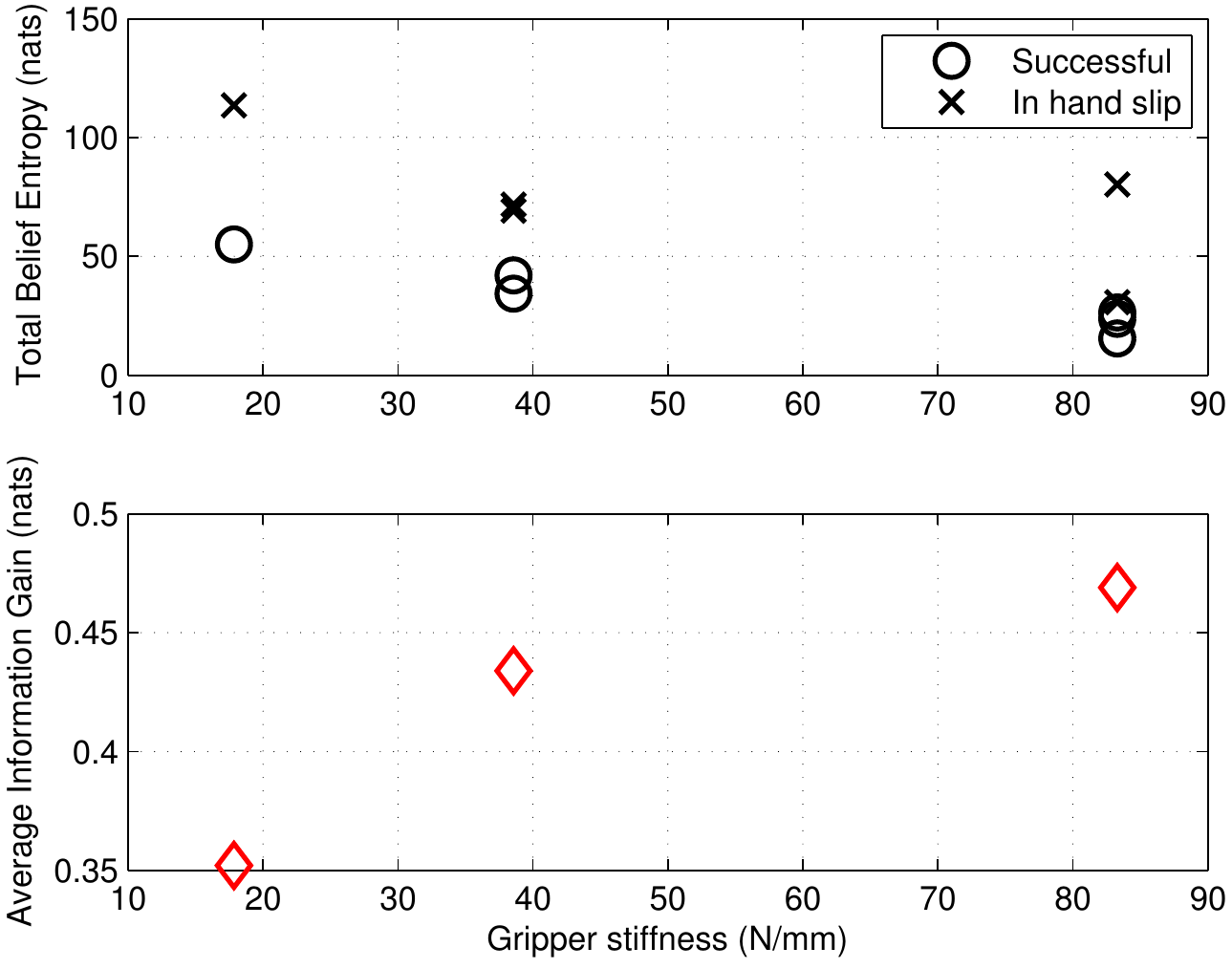}
\caption{Total belief entropy over compliance at the gripper.  For both successful and in hand slip trials, the uncertainty increases as the compliance decreases.  \label{fig:entropy_in_monitoring}}
    \vspace{-4mm}
\end{figure}

A typical mode detection from a trial with in-hand slip can be seen in Figure \ref{fig:entropy_detail}, where the belief entropy (i.e. mode uncertainty) increases substantially during the transitions between the modes. Comparing the total entropy over a number of experiment iterations, we find that decreasing the gripper stiffness increases the total uncertainty, as seen in Figure \ref{fig:entropy_in_monitoring}. 

The information gain is calculated using the fully-observed information gain \eqref{info_gain_full} with the gripper stiffness as $K_1$, and the average taken over $|q^1-q^z2|\in [0, 0.5]$ mm with a flat prior $b_n = 1/N$. It can be seen that the higher stiffness has higher information gain, corresponding with lower total belief entropy.

\subsection{Collision detection - High Payload Robot}
\begin{figure}
\centering
\includegraphics[width = 0.8\columnwidth]{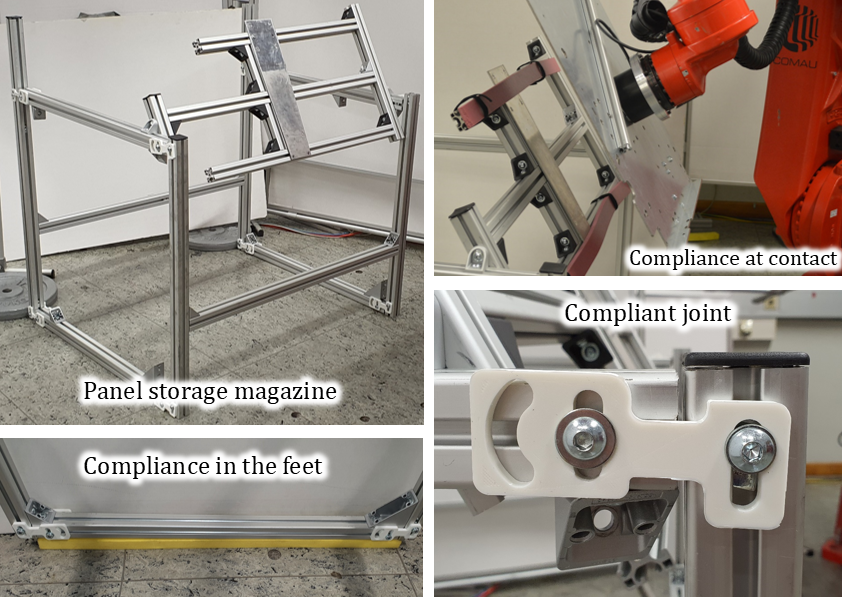}
\caption{Contact magazine with the various locations of compliance, using strips of viscoelastic foam, or 3d printed flexure joints. } \label{magazine_setup}
    \vspace{-4mm}
\end{figure}

A 16 Kg plate is mounted to a large industrial robot as seen in Figure \ref{magazine_setup}, and brought into contact with a compliant magazine, where compliance is introduced in one of three locations: (i) in the joints of the magazine, (ii) on the contact surface, and (iii) under the feet of the magazine.  The robot was driven into contact in using admittance control with a desired contact force of 30 N. Pure admittance control is used to collect data for model identification, then contact detection is activated, where a simple force threshold of $6$ N is used (gravitational forces of the plate are compensated).

\begin{table*}
\centering
\begin{tabular}{ r|c|c|c|c|c}
 Condition & $K_1$ & $M_2$ & $B_2$ & $K_2$ & $\underline{\Delta}_I$   \\
 \hline
 Flex Joints & 17.4e4 & 20.1 & 305 & 2630 & 0.131  \\ 
 Compliant Surface & 1.3e4 & 63.9 & 1870 & 7350 & 0.0743 \\
 Compliant Feet & 2.61e4 & 69.3 & 1080 & 1.81e4 & 0.348 \\ \hline \hline
 Flex Joints Gradient & 7.7e-6 & 7.6e-3 & 2.2e-6 & -4.8e-10 & - \\ 
 Comp Surf Gradient & 1.5e-5 & 1.8e-3 & 1.7e-6 & 1.3e-10 & -\\ 
 Comp Feet Gradient & 2.5e-5 & 0.8e-3 & 1.3e-6 & 4.8e-10 & -\\
 
\end{tabular}
\caption{Identified dynamic parameters for the magazine contact  \label{magazine_parameters}}
\end{table*}
\begin{figure}
\centering
\includegraphics[width = 0.8\columnwidth]{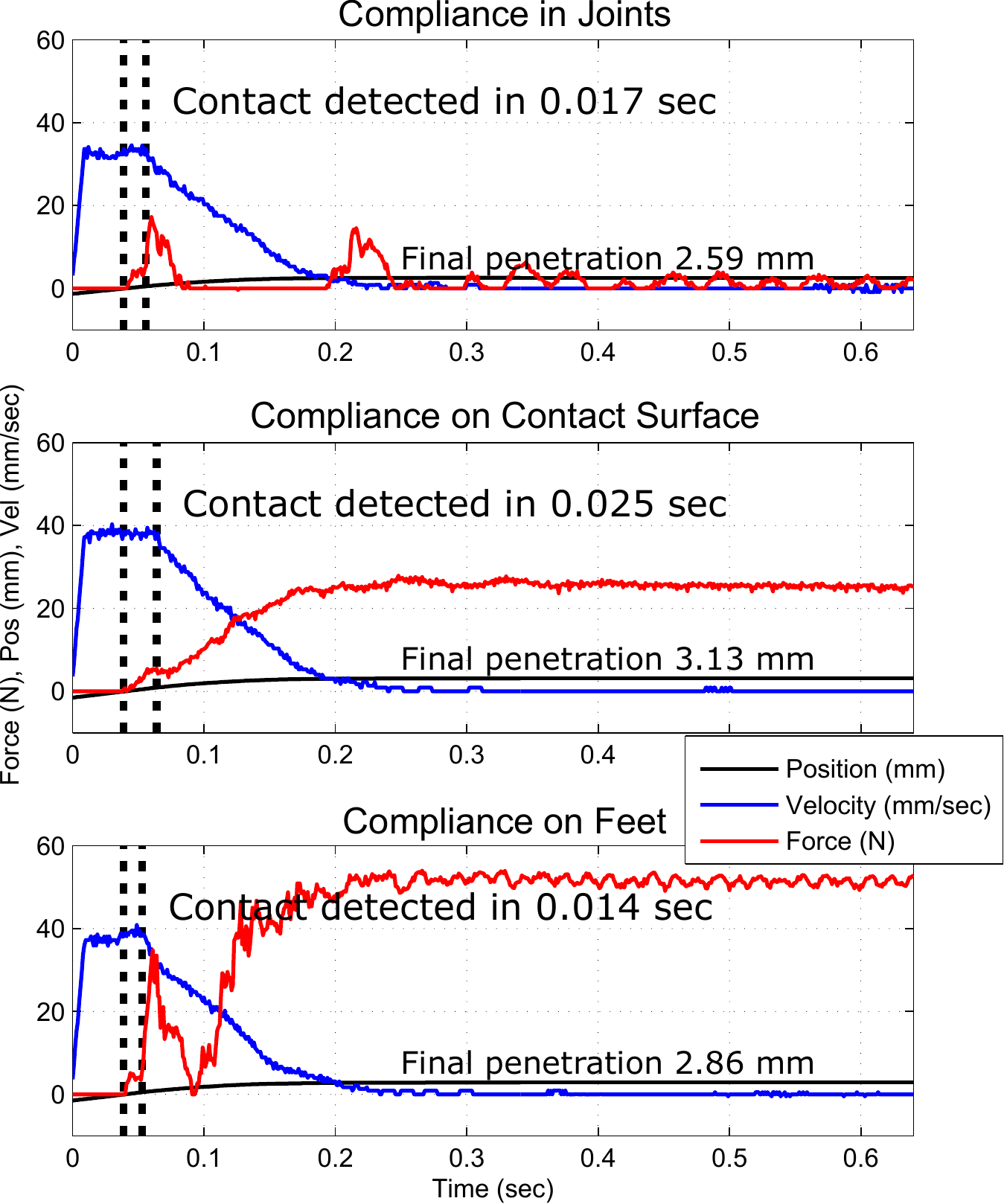}
\caption{Contact force, position, and velocity in experiments with the contact detection \label{magazine_contact_comparison}}
    \vspace{-4mm}
\end{figure}

As seen in Figure \ref{magazine_contact_comparison}, the fastest contact detection occurred with compliance on the feet (bottom), where the hard contact and larger effective magazine inertia results in an initial peak in force.  Compliance in the joints results in the 2nd fastest contact detection -- note that in this case contact was lost as the magazine springs back with the lower $K_2$. The slowest contact detection was with compliance at the contact surface, where force increases slowly. 

\begin{figure}
\centering
\includegraphics[width = 0.7\columnwidth]{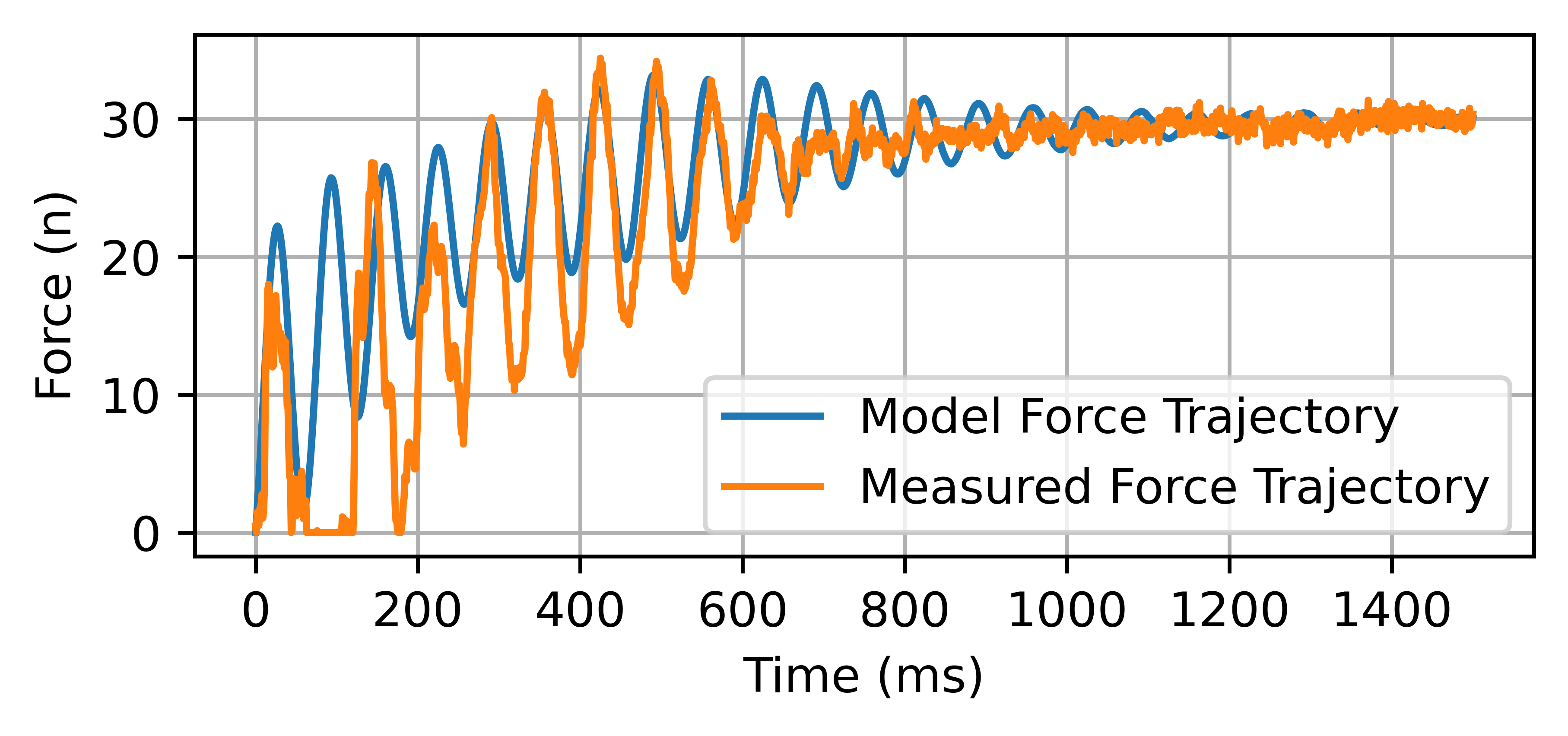} \\
\includegraphics[width = 0.7\columnwidth]{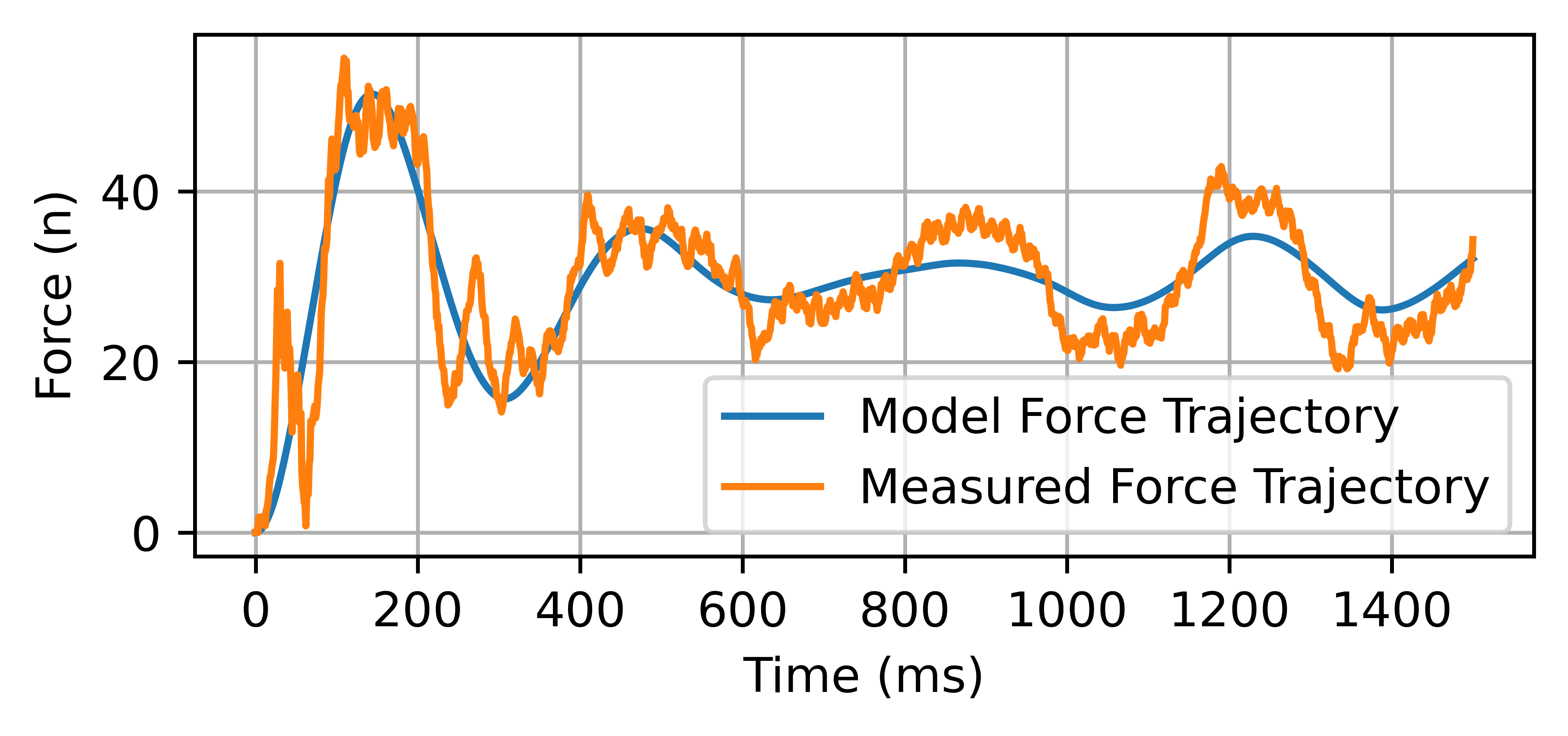}
\caption{Identified model and real response, for flex joint magazine (top) and compliance in the feet (bottom). Contact is made with force control, with desired force of 30 N. Note that admittance parameters are set poorly for the bottom contact, but the identified model is still accurate. \label{magazine_model_comparison}}
\vspace{-4mm}
\end{figure}

A model for these three compliance cases is then identified following the inertial model in Figure \ref{two_mass}, where the measured robot position is $q^1$, unmeasured magazine position is $q^2$ and the measured force is that on $K_1$. These parameters are identified by doing a least-squares fit $\min_{K_1, M_2, B_2, K_2} \sum_t\Vert f^m_t - f_t\Vert$, where $f^m_t$ is the model output given the observed $q^1_t$ as input, and can be seen in Table \ref{magazine_parameters}, with time plots in Figure \ref{magazine_model_comparison}.

From these models, the $\underline{\Delta}_I$ can be calculated in the partially observed case with $q_t$ at time $t=0.002$. It can be seen in Table \ref{magazine_parameters} that the compliant feet has the highest information gain, followed by the compliant joints and compliant surface, corresponding to the contact detection performance. Because the $\underline{\Delta}_I$ calculation is composed of differentiable operations, the gradient can be numerically calculated with an automatic differentiation toolbox \cite{andersson2019}, also seen in Table \ref{magazine_parameters}.  It can be seen that increasing $K_1$ and $M_2$ make the largest increase in $\underline{\Delta}_I$, while $B_2$ and $K_2$ play a more minor role. The gradient validates an experimental observation: a higher contact inertia $M_2$ helps improve contact detection. 

\subsection{Collision Detection - Medium Payload Robot}
\begin{figure}
\centering
\includegraphics[width = 0.8\columnwidth]{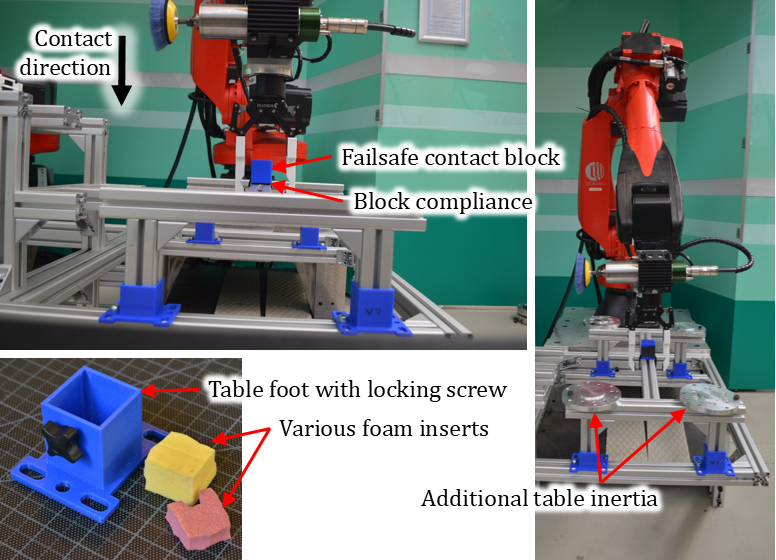}
\caption{Compliant table with 7 Kg payload robot, where compliance can be adjusted in the feet of the table or between the failsafe block and table.\label{table_setup} }
\vspace{-4mm}
\end{figure}

\begin{figure}
\centering
\includegraphics[width = 0.8\columnwidth]{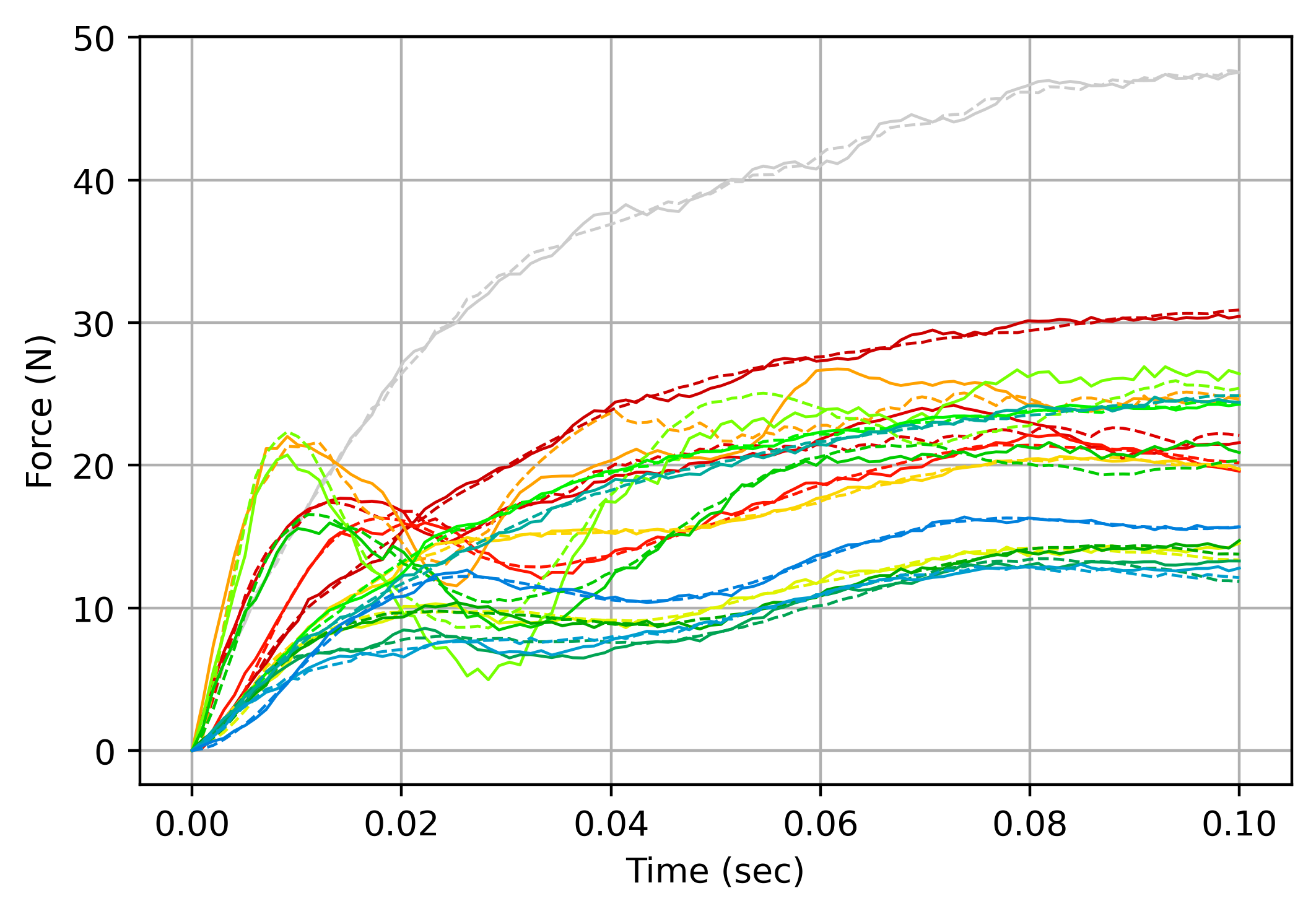}
\caption{Traces of contact force (solid line) vs identified model response (dotted). Detailed responses can be better seen on the online repository, these responses show the range of force profiles and model fit. } \label{table_traces}
\vspace{-4mm}
\end{figure}

A compliant table, seen in Figure \ref{table_setup}, was used to investigate the impact of location and quantity of compliance on the ability to detect contact. By varying the compliance in the feet and contact surface, as well as the inertia of the table, a range of contact responses were achieved as seen in Figure \ref{table_traces}. A simple contact detection threshold of $6$ N was applied, and the performance can be seen in the attached video. Model identification was done as above, but a two-inertia model was found to provide better model fit, requiring a straightforward extension to \eqref{two_mass_dyn} which is omitted due to space constraints but in the published code.

\begin{figure}
\centering
\includegraphics[width = 0.8\columnwidth]{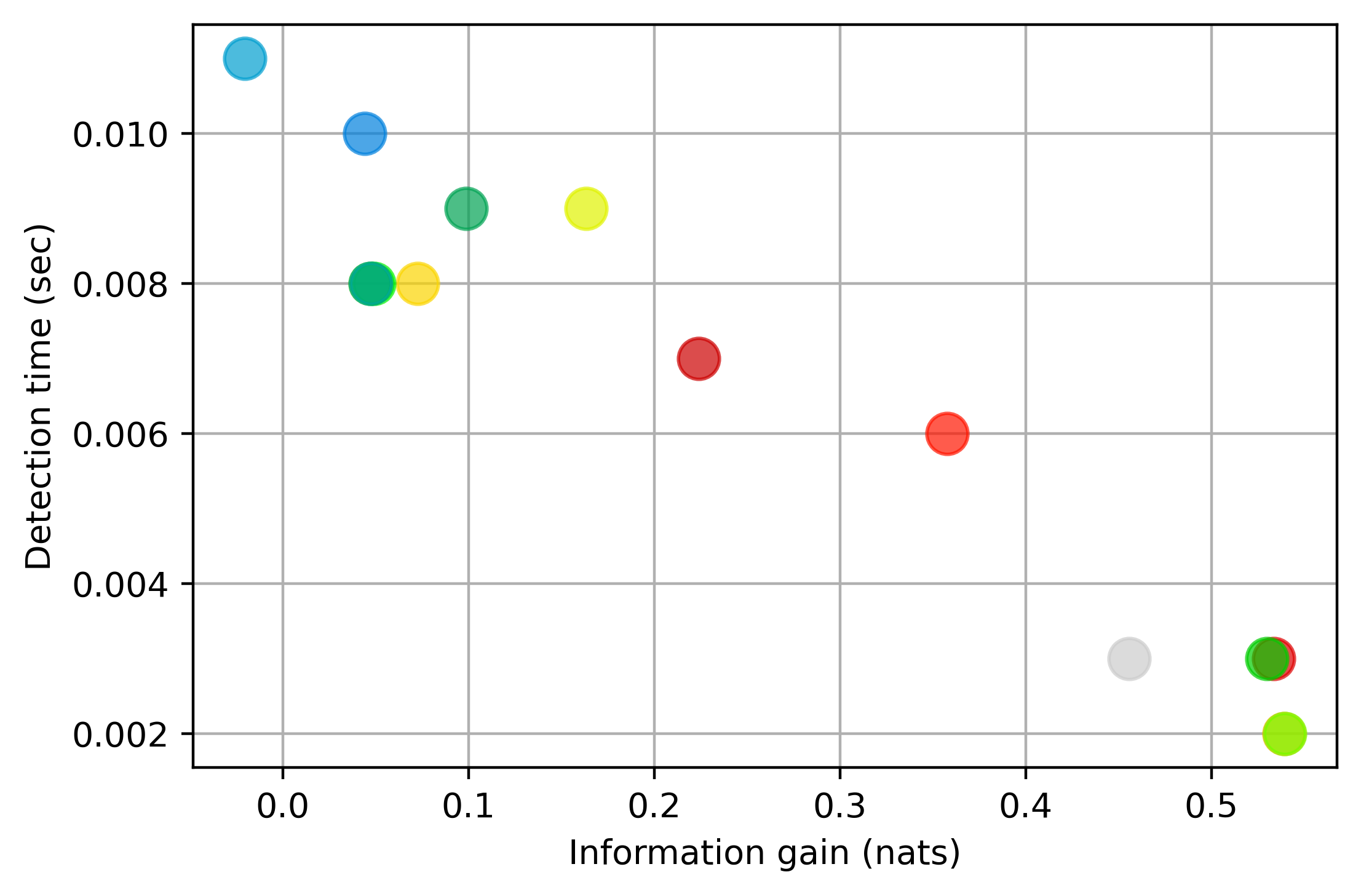}
\caption{Information gain vs time to detect contact, colors are consistent with those used in Figure \ref{table_traces}.  As information gain increases, the time required to detect contact decreases. Hard contact and higher table inertia gave the fastest detection. } \label{table_ig}
\vspace{-4mm}
\end{figure}

The information gain, calculated with \eqref{partially_obs} at the state $t=0.002$ after contact, is compared with the time taken to detect the contact in Figure \ref{table_ig}.  It can be seen that as information gain increases, time to detect contact decreases. 

The configurations where contact compliance was removed, and where additional weight was added to the table, had higher information gain and a higher contact velocity could be safely achieved - further detailed plots are available online with the experimental data.

\section{Conclusion}
The impact of mechatronic design, especially compliance, on the ability to detect contact conditions was presented, and formalized in a metric of the information gain. This metric is validated to predict the certainty with a complex assembly task can be monitored, and the speed with which collision can be detected. 

In the case of introducing compliance to the environment, it is shown that this compliance is best introduced not at the point of contact, but deeper into the kinematic structure.  This allowed faster contact detection in both the magazine and table contact tasks.


\bibliographystyle{IEEEtran}
\bibliography{lib}

\end{document}